# Localizing the Object Contact through Matching Tactile Features with Visual Map

Shan Luo[1], Wenxuan Mou[2], Kaspar Althoefer[1], and Hongbin Liu[1,*]

*Abstract*—This paper presents a novel framework for integration of vision and tactile sensing by localizing tactile readings in a visual object map. Intuitively, there are some correspondences, e.g., prominent features, between visual and tactile object identification. To apply it in robotics, we propose to localize tactile readings in visual images by sharing same sets of feature descriptors through two sensing modalities. It is then treated as a probabilistic estimation problem solved in a framework of recursive Bayesian filtering. Feature-based measurement model and Gaussian based motion model are thus built. In our tests, a tactile array sensor is utilized to generate tactile images during interaction with objects and the results have proven the feasibility of our proposed framework.

## I. INTRODUCTION

Eyes and hands are both capable assistants for us humans and they two always coordinate to fulfil complex tasks, e.g., grasping, exploration and object recognition, etc. Accordingly, vision and the tactile sensations of hands are synthesized during our perception of the ambient world. When we intend to grasp and/or manipulate objects with our hands, we tend to fixate on a target to have a glance at it first with our eyes to get some key features, i.e., corners, edges, curves, textures and key points. While once we grasp it, these distinct features become unobservable because vision is occluded by the hand. Therefore, vision becomes no more sufficient, especially when manipulated objects are smaller than the hand. In these cases, touch sensation of our finger pads or the palm can assist us to make up this information loss: corresponding features are sensed in the tactile modality; positions and poses of objects in our hands can therefore be inferred. By tracking and matching these clues through vision and tactual sensation, our hands can be adapted swiftly and perform grasps flexibly. Similarly, during exploration in unknown environments, eyes direct the movement of hands to targets in general situations whereas conversely fingers act as "eyes" to perceive object properties in dark situations or when we fumble in narrow places. Furthermore, for object recognition, eyes provide initial information of the object, e.g., its size and general shape, while tactile discrimination supplements these features with texture and more detailed shape. From all these examples, it can be noted that prominent features serve as a bridge between the vision and tactile sensing, making the hand movements consequent.

[1]Shan Luo, Kaspar Althoefer, Hongbin Liu are with the Centre for Robotics Research, Department of Informatics, King's College London, WC2R 2LS, UK (e-mail: shan.luo@kcl.ac.uk, k.althoefer@kcl.ac.uk, hongbin.liu@kcl.ac.uk).

[2]Wenxuan Mou is with School of Electronic Engineering and Computer Science, Queen Mary University of London, London E1 4NS, UK (email: w.mou@ qmul.ac.uk).

[*]This indicates the corresponding author.

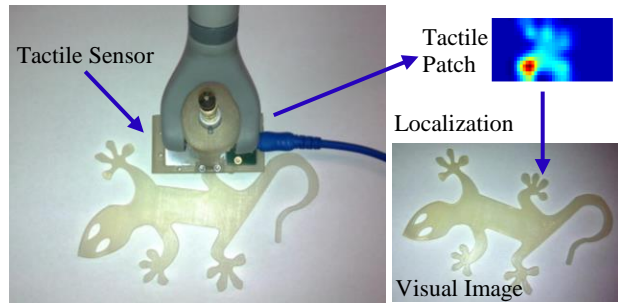

Fig. 1. Left: Experimental setup, which consists of a webcam (not shown here) and a tactile sensor attached to a manipulator. The test object is selected as a 3D print gecko model. Right: Localization flow to localize the tactile reading in the visual map/image of the object.

The aforementioned mechanisms have been partly proved in neuroscience and psychophysics. Thanks to the functional Magnetic Resonance Imaging (fMRI), researchers have found that certain regions of human brain cortex are shared by both visually and haptically originated representations of objects. Zangaladze *et al.* [1] confirmed the truth that visual cortical processing is involved in normal tactile perception, i.e., discrimination of orientations in this case. Amedi *et al.* [2] gathered evidence that indicates vision and touch indeed share the same shape representations in the cortex, which bounds vision and tactile modalities together to generate a coherent percept. Though the exact level of bimodal shape representation (basic features level or holistic object shape level) in cortex still has not been revealed until now, it can be concluded that visual and haptic inputs share certain processing of object representations in our brain. To employ this idea in robotics, we propose to use same feature descriptors in both vision and tactile sensing to localize the tactile features in visual images. Fig. 1 depicts our experimental system: a webcam (not shown in the figure) and a tactile sensor are utilized to obtain visual and tactile images of objects respectively; a Phantom Omni haptic device to which the tactile sensor is attached is employed to acquire the positions of the tactile sensor. The remaining paper is organized as follows. The related research is reviewed in Section II, including current methods to fuse vision and tactile sensing, the state-of-art in shape recognition based on tactile images and robotic localization methods. The approach to localize tactile readings in visual images is introduced in Section III. A set of experiments are carried out to evaluate our method, as described Section IV. Finally, the paper is concluded in the last section.

## II. RELATED WORK

### A. Integration of vision and tactile sensing

In robotics, the attempts to fuse vision and touch to recognize and represent objects can be dated back to decades ago. Most of the researchers only take tactile sensors as devices to verify contacts due to their low resolution. For instance, Allen [3] supplemented object surfaces created by 3D vision with simple contact sensing to confirm object geometry profiles. However, due to low resolution of the tactile sensor, it was only used to verify specific features, i.e., holes, cavities, and obtain relational constraints of these features. Yeung *et al.* [4] also assisted vision with haptic exploration to determine 3D positional parameters in relatively structured environments, and tactile sensing performed a similar role. Recently, Ilonen *et al.* [5] proposed an optimal estimation method to learn object models during grasping via visual and tactile data by using iterative extended Kalman filter (EKF): visual features are first collected to create an initial hypothesis, then tactile readings are gradually added to refine the object model. A similar work was carried out by Bjorkman *et al.* [6] but Gaussian process regression was utilized instead. However, in both of their work, tactile measurements were still used to test whether and how the robotic hand is in contact with the object. Some other researchers attempted to combine visual and haptic information to estimate the pose of an object [7]. However, to the best of our knowledge, there still has no work been done to take both visual and tactile readings as images and utilize the same descriptors to extract features through two modalities for object contact estimation.

*B. Tactile image based shape recognition*

Thanks to the increasing spatial resolution and spatiotemporal response in last few decades [8], tactile sensors have demonstrated the ability to serve as an "imaging" device. Schneider *et al.* [9] took tactile images as features directly to recognize objects in a framework of Bag-of-Words (BoW) originated from computer vision. In the same framework, Pezzementi *et al.* [10] took one step further: multiple kinds of features were extracted from tactile readings and their performances were compared. And they also proposed a mosaic method to synthesize local geometric surfaces obtained from tactile images to recover the object-level geometric shape using histogram and particle filters [11], in which the objects were a set of raised letters. Liu *et al.* [12] recognized objects by classifying local features through the covariance analysis of pressure values in tactile images. The authors also contributed a method to recognize contact shapes by using neural network [13]. In [14], moment analysis and principal component analysis were used to create low-dimensional features for tactile contact patterns. All of these works show the potential to integrate tactile readings with visual images.

*C. Robotic localization in a map*

The robotic localization problem has been known as simultaneous localization and mapping (SLAM) and many researchers have put their effort into it. Until now, most of the work has been done with visual inputs [15] whereas only a few researchers explored to localize the robot with the tactile information and single tactile modality was used in their work [16]–[18]. A "haptic map" was created from the tactile measurements during the training and it is used to localize features (bump, snap and grommet in their case) embedded in flexible materials during robot manipulation [16]. Fox *et al.*[17] introduced the grid based SLAM to robotic navigation with biomimetic whisker sensors by deriving timing information from contacts and a given map about edges in a small arena. Li *et al.* proposed to localize tactile images with a height map via image registration to help localize objects in hand [18]. However, there is still no work present to demonstrate localizing tactile features in a visual map.

Compared to the previous work, our contributions can be summarized as:

*1) A framework to localize tactile readings in visual images using recursive Bayesian filtering is proposed and verified.*
*2) Feature descriptors of the same type are first extracted for both visual and tactile images.*
*3) A novel approach is provided that is promising to facilitate robotic grasping and other manipulations in hand, by integrating visual and tactile information.*

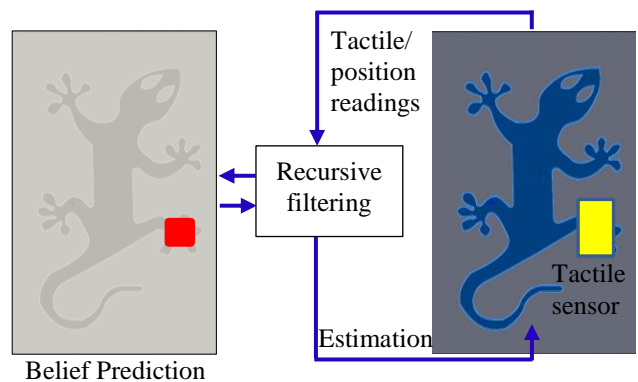

Fig. 2. Robot object interaction for localization and the data flow. Left: prediction of the contact location in the visual map, with the red region as the most probable contact location. Right: control and measurement update.

III. METHODOLOGY

*A. Overview*

To perform the manipulations in hand consistently, it is important, but also tricky, to build up connections between the tactile and visual sensations as the robotic hands always occlude the vision. Thus it is aimed to find correspondences between vision and tactile sensing. The visual image of the object is treated as a map and it can be obtained prior to hand operations. As tactile exploration is carried out, the relative location of the robotic finger to the object, which cannot be sensed directly, is inferred in the light of the tactile measurements and the movements of the finger. Based on this pipeline, the problem is viewed as a matter of estimating the location of a robot in a map; thus a recursive Bayesian estimation process from the mobile robotics is employed. Following the notations in [19], the problem is restated as follows. To acquire the information about the location (state) $x_t$ of the tactile sensor in the world $m$ (visual image) at each time step $t$, the tactile readings $z_t$ (sensor measurements) and movements of the tactile sensor $u_t$ in the 3D space (control actions) are collected to calculate the belief distributions over possible locations (states). For simplicity, there are a set of assumptions: both the robot base coordinate and the object keep fixed during the exploration, which makes the map static; the tactile sensor possesses a planar surface and the sensing elements are evenly distributed, which enables it to act as a tactile "camera"; the explored object has a surface with distinct 2D features in a plane, perpendicular to which the tactile sensor is as it interacts with the surface; and the exploration process is assumed to take place in a hidden Markov Model (HMM).

Fig. 2 depicts the tactile sensor-object interaction for localization. As the tactile sensor explores the object surface, tactile images are acquired from each interaction and its positons in the world coordinates can also be obtained from forward kinematics of the robot. However, due to corruption by noise of the measurements or unmodeled exogenous effects, the relative location of the tactile sensor in the object map cannot be calculated in a deterministic form. Thus a recursive filtering process is taken to compute its belief distributions over possible locations in the object map based on the measurements and control data, which results into two steps shown in Table 1. The input for calculating the belief $bel(x_t)$ at time $t$ is the belief $bel(x_{t-1})$ at time $t$-$1$, new control action $u_t$ and new measurement $z_t$. The *control update* is first carried out to get the belief $\overline{bel}(x_t)$, only incorporating the control $u_t$. $\overline{bel}(x_t)$ is obtained by the integral (sum in discrete case) of the product of the prior $bel(x_{t-1})$ and the probability that the control $u_t$ induces a transition from $x_{t-1}$ to $x_t$. Then the measurement $z_t$ is involved to calculate the final belief $bel(x_t)$, which is called *measurement update*. The belief $\overline{bel}(x_t)$ from last step is multiplied by probability that the measurement $z_t$ may have been observed for each hypothetical posterior state $x_t$. As the resulting product is generally not a probability, the result is normalized to be integrated to 1, with the normalization constant $\eta$. Besides, to calculate the posterior belief recursively, an initial belief $bel(x_0)$ at time t=0 is required as the boundary condition. In our case, we assume there is no knowledge acquired before the exploration; therefore, the probabilities are evenly distributed for all the locations in the object map.

*B. Motion model*

As described in [19], there are two probabilistic motion models $p(x_t \mid u_t, x_{t-1})$: one is acquired from the velocity data whereas the other is derived from odometer information. In our case, the localization device takes a role of odometer: the world coordinates of the center of the tactile sensor at each point are recorded and thus the travelled distance and turned angle at each time step can be computed. And the state transition distribution is modeled by a locally-weighted Gaussian distribution defined with respect to the nearest $k$ states ($k$=8 in our case) to $x_t$, measured in Euclidean distance.

$$p(x_t|x_{t-1}, u_t) \sim N(x_t; \hat{x}_t, \sum x_t),$$

where $N(x; \mu, \Sigma)$ denotes the Gaussian probability density function over $x$ with mean $\mu$, and covariance $\Sigma$.

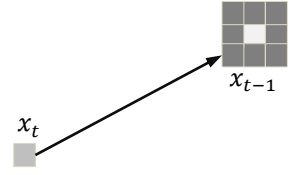

Fig. 3. Motion model in the image map. The motion of the tactile sensor is noisy, modeled in Gaussian distribution. In the estimation of state $x_{t-1}$, the brighter the pixel is, the more probability is assigned to that the sensor moves to this location at time *t*.

*C. Feature-based measurement models*

The normalized cross correlation method is first attempted to calculate the distance between each tactile image and each sliding window in the visual image. However, due to its poor performance, we turned to use more advanced and richer feature descriptors to compute the distance. The detailed measurement model is described as follows.

There are already multiple feature descriptors created to represent visual images as discussed in [20]. However, there are still no feature descriptors tested by both visual and tactile images. Taking the invariance to translation and rotation into consideration, it is proposed to adapt the widely used Scale Invariant Feature Transform (SIFT) descriptors [21] in computer vision to both scenarios. To simplify the problem, it is assumed that the transformation between the visual images and objects is already known, which means that the real dimension and shape of the object can be recovered from the image. It is a reasonable assumption as the development of depth sensor cameras and consequent processing already makes it a reality. For tactile sensing, the real dimension and shape of the interacted object can be mapped to the tactile sensor directly; thus there is no transformation needed for tactile images. In classic SIFT algorithms, key points, e.g., corners, are first detected and a multitude of distinctive features can be obtained in one image due to affluent information. However, there is scant information and much less such features can be extracted from each tactile image. Based on these ground truths, the procedures of scale-space pyramids and key point localization are removed as in [22]. And to make it more robust, each tactile image is segmented into three overlapped sub-images of the same size and one SIFT descriptor is extracted from each sub-image, taking sub-image centers as "key points". More details can be found in our previous work [23]. Different from classic SIFT descriptors with a dimension of 128, descriptors of 32 elements are obtained instead in our case by reducing sampling areas from a 4×4 grid to a 2×2 grid, to minimize the computation time. However, there is no significant performance deterioration observed in our experiments, compared with the classic 128-element SIFT descriptors.

The same approach is also applied to the visual images, extracting three SIFT descriptors from each sub-image. A sliding window with a size of the tactile image is carried out to get the matching probability for each pixel. As shown in Fig. 4, three SIFT descriptors $f_{i,j,m}$ are obtained from the region starting with the pixel $p_{i,j}$. And the distance between this sub-image and the tactile image $z_t$ at time *t* is gained by calculating the Euclidean distances between these three

---

**Algorithm Bayes_filter**($bel(x_{t-1})$, $u_t$, $z_t$):
*for all $x_t$ do*
  $\overline{bel}(x_t) = \int p(x_t|u_t, x_{t-1})bel(x_{t-1})dx_{t-1}$
  $bel(x_t) = \eta p(z_t|x_t)\overline{bel}(x_t)$
*endfor*
*return $bel(x_t)$*

Table 1 The Bayesian filtering framework

descriptors and features $f_{t,n}$ extracted from the tactile reading $z_t$ as in (1).

$$d(p_{i,j}, z_t) = \sum_{m=1}^{3} \sum_{n=1}^{3} \sum_{k=1}^{32} (f_{i,j,m}^k - f_{t,n}^k) \quad (1)$$

The probability $p(z_t|x_{i,j,t})$ that the measurements $z_t$ at time step $t$ can be induced at state $x_{i,j}$ is calculated in (2). The parameter $\eta'$ is the normalization factor, similar to the one in the Table 1. The posterior probability $bel(x_t)$ can thus be obtained.

$$p(z_t|x_{i,j,t}) = \frac{\eta'}{d(p_{i,j}, z_t)} \quad (2)$$

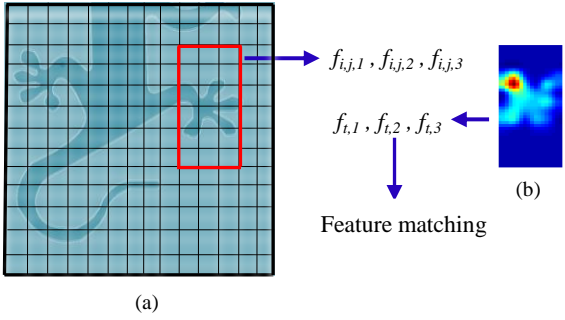

Fig. 4. Feature matching between visual and tactile images. (a) A sliding window (marked with red rectangle outline) in the visual image, with a start point marked red. (b) Tactile image $z_t$ at time $t$.

## IV. EXPERIMENTS AND RESULTS

### A. Experimental setup

The experimental setup comprises three parts: a webcam, a tactile sensor and a Phantom Omni device. The image from the webcam is resampled into a resolution of 120×120. The resistive tactile array sensor is from Weiss Robotics and it consists of 84 sensing elements evenly distributed in 14 rows and 6 columns. It has a size of 51 mm ×24 mm as a whole and 3.4 mm square for each cell. The sensor is covered by elastic rubber foam to conduct the externally applied force. The Phantom Omni device serves as a robotic manipulator, to which the tactile sensor is attached. And the position of the stylus tip is obtained by the forward kinematics based on the robot base frame, which is taken as the position of the tactile sensor center.

To test our algorithm, objects with distinct features in a plane are first used in elementary experiments. As a first trial, we took a pair of scissors as our experimental object. The good localization performance proves the feasibility of our method, but due to the limited space, the results are not present here. A 3D printed gecko model with more features is utilized to further test our algorithm, illustrated in Fig. 1. The gecko has a 2D shape that protrudes from the base 4mm. The tactile sensor is controlled to press it at multiple steps to follow the surface of the gecko. During each interaction the sensor plane is kept normal to the surface of the gecko.

The raw readings are preprocessed in two steps: 1). if in a tactile image the maximum value is lower than the specific threshold or the sum of all elements is smaller than a predefined decision value, it is considered as collected unintentionally and deleted. 2). the readings are then normalized to the maximum value of each tactile image, hence, falling into [0, 1]. The tactile images are resized to the scale of the visual image, 42×18 in our case. As described in Section III C, three descriptors are extracted from one tactile image, with a sample sub-patch size of 18 and a grid spacing of 9. A sliding window is used to match the segmented sub-images with the input tactile readings and the obtained probability is assigned to the starting pixel of each sub-image. But the use of this method brings a significant computational burden, therefore, parallel processing is employed to boost the calculation: a set of the sub-images are taken to compute their distances to the input tactile image at the same time.

### B. Results and analysis

An example exploration process is illustrated in Fig. 5, taken from the right to the left and from the bottom to the top of the gecko. The process is divided into six steps thus results at seven locations are present, with the first column at the initial position. The tactile sensor stays at each location for a certain time and 20 readings are thus collected at each location. In the first row of Fig. 5, sample tactile images at each step are illustrated. It can be noticed that even though the resolution of the tactile images is still low, certain features can be observed in each image. The corresponding ground truth locations of the tactile sensor are shown in the second row. And the probabilities of the tactile sensor at different states are illustrated in the third row.

Before the exploration process, the probabilities are evenly distributed over all the possible states (not shown in Fig. 5). However, once the exploration process starts, the sates occupied by the gecko show larger possibilities over the other states, as shown in the first probability map in the third row. But due to the inaccurate feature matching, it is hard to determine which part of the gecko being interacted. Along with the exploration process, the probabilities converge. Every time one new measurement is taken, the more certain the robot can determine the location of the pressed part. However, the movements of the tactile sensor deteriorate the tendency at the same time. At the end of the exploration process, the robot can locate the tactile sensor in the gecko map with a large certainty. The localization performance is calculated by the difference between the estimated most probable location of the tactile sensor and the ground truth locations, as shown in Fig. 6. It can be noticed that the localization errors tend to decrease through the localization process.

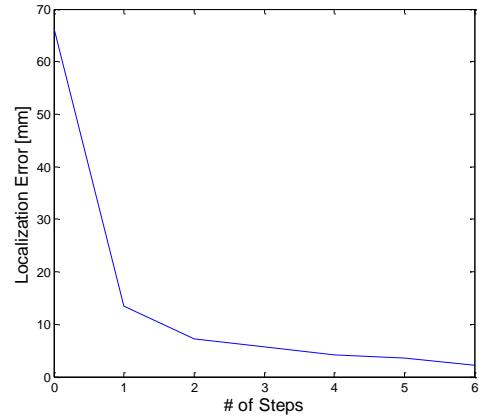

Fig. 6. Localization performance with respect to number (#) of steps of touch, and the localization errors are found to decrease through the whole process.

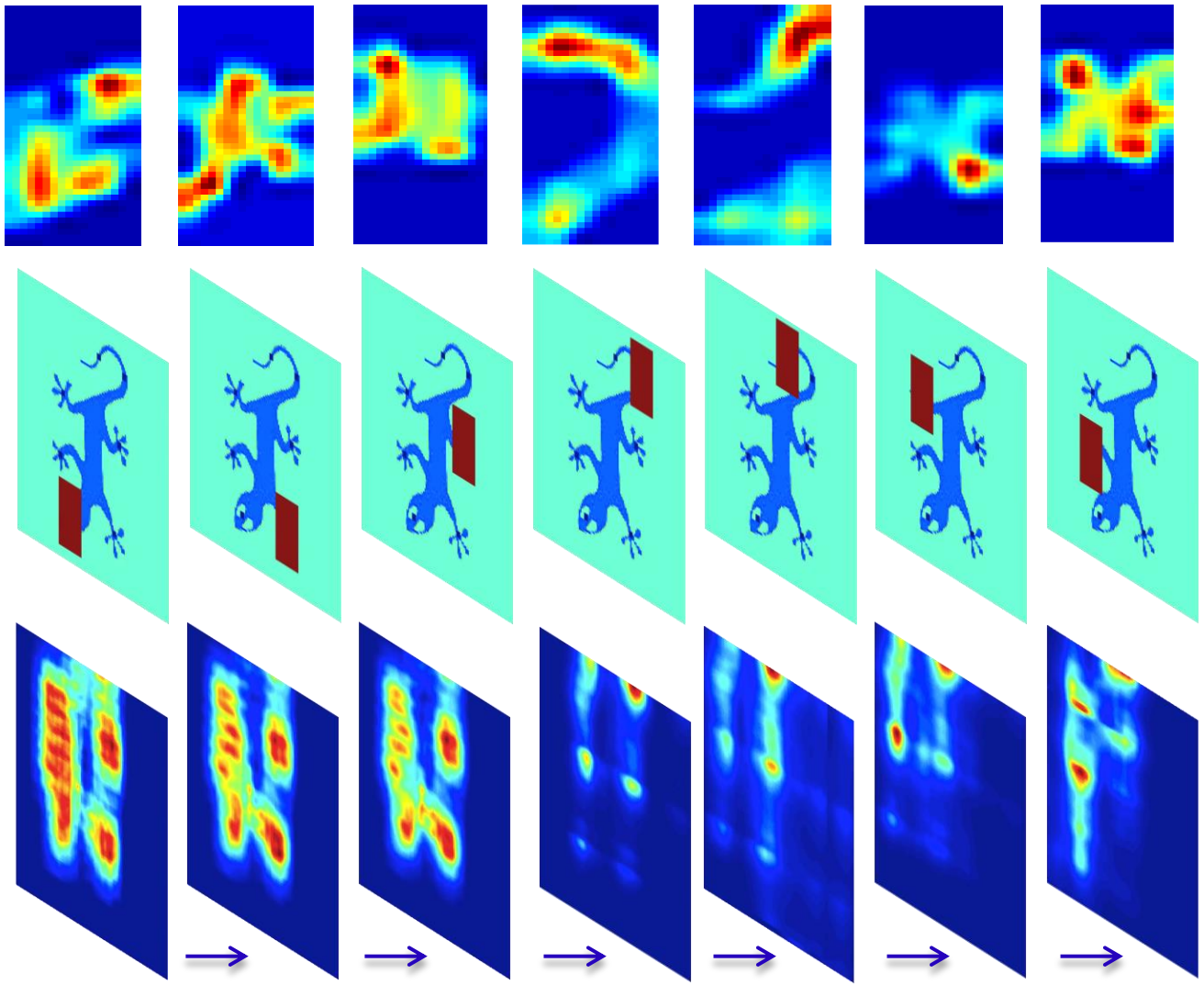

Fig. 5. Sample localization process and results. First row: obtained tactile images at each step; Second row: corresponding ground truth locations of the tactile sensor (marked as red) in the visual map of the gecko. Last row: Corresponding probability distribution of locating the tactile sensor at different states.

In total, 22 experiments are carried out in a similar manner, but the tactile sensor explores the gecko model in different exploration paths. In other words, the tactile sensor interacts the gecko at the same locations as the ground truths shown in the second row in Fig. 5, but in different orders. In most of them, a good localization performance has been achieved in limited six steps and the robot can locate itself with an error of less than 5 mm. There are only two exceptions in which the robot locates itself with a large error through the whole process. The probable reason for how it happens is that the influence of large rotation and translation noises counterweighs the benefits of the new input tactile readings. The histogram distributions of datasets with different localization performance are illustrated in the Fig. 7. Among the datasets in which good performance can be achieved (the robot can locate itself within 6 steps), the number of steps to achieve satisfactory localization results is also counted, as shown in Fig. 8. It can be found that the robot can mostly localize itself within 4 steps. To conclude, the experimental results prove the feasibility of our proposed framework in localizing the tactile sensor in a visual object map.

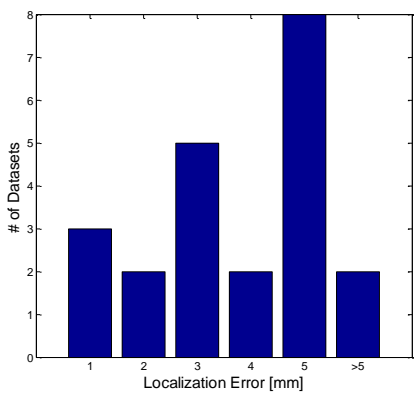

Fig. 7. Number (#) of datasets with different localization errors, it can be noted that in most trials a good localization can be achieved within six steps.

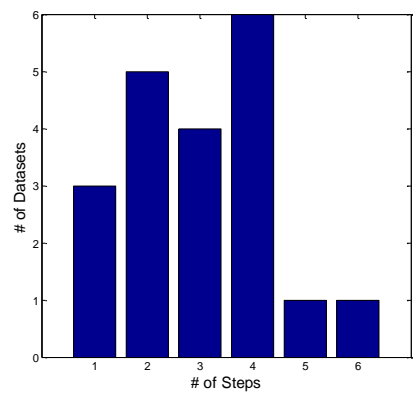

Fig. 8. Number (#) of datasets with different steps to achieve good localization performance, it can be noted that the robot can mostly localize itself within 4 steps.

## V. Conclusion and future work

This paper proposes a novel approach to integrate the vision and tactile sensing via localizing the tactile readings in the visual object image, which is viewed as a matter of locating a robot in a map. The recursive Bayesian filtering is employed to estimate the belief distributions over all the possible locations in the visual image. The movements of the tactile sensor are treated as odometer readings and the Gaussian noises are used to model the motion. In measurement update, revised SIFT descriptors are extracted from both the tactile and visual images and the belief distributions are updated by feature matching. The algorithm is tested by locating a moving tactile sensor in a visual object map and it is proved that the relative location can be inferred with a recursive filtering process. It provides a promising approach to facilitate robotic grasping and hand manipulations by integrating both visual and tactile information.

Due to the early stage of this research, there are several directions for future work. In this paper, a Phantom Omni is used as a manipulator whereas its dimension constraints its movement space. Thus a robotic hand will be involved to test wider applications of our method. As the high computational SIFT descriptors are employed here, it cannot be applied to the swift manipulations in hand directly thus low dimensional descriptors can be attempted and used instead in future. Besides, the object is fixed in the robot base coordinate system in our case but in real scenarios objects are always translated and rotated by the hand. It induces more noises to the motion model, which is worth being further probed in future work.

## VI. Acknowledgments

This work was partially supported by the King's-China Scholarship Council.